\pdfoutput=1
\documentclass[11pt]{article}

\usepackage{emnlp2021}

\usepackage{times}
\usepackage{latexsym}
\usepackage{multirow}
\usepackage{graphicx}
\usepackage{graphics}
\usepackage{multicol}
\usepackage{xcolor}
\usepackage{verbatim} 
\usepackage{url}
\usepackage[utf8]{inputenc}
\usepackage{enumitem}
\usepackage{array}
\usepackage{amsmath}
\usepackage{makecell}

\usepackage[T1]{fontenc}
\usepackage[utf8]{inputenc}

\usepackage{microtype}

\title{HintedBT: Augmenting Back-Translation with Quality and Transliteration Hints}

\author{Sahana Ramnath, Melvin Johnson\thanks{*equal contribution}, Abhirut Gupta\footnotemark[1], Aravindan Raghuveer \\ Google Research \\ \texttt{\{sahanaramnath, melvinp, abhirut, araghuveer\}@google.com}}
\begin{document}
\maketitle
\begin{abstract}
Back-translation (BT) of target monolingual corpora is a widely used data augmentation strategy for neural machine translation (NMT), especially for low-resource language pairs.  To improve effectiveness of the available BT data, we introduce HintedBT---a family of techniques which provides hints (through tags) to the encoder and decoder. First, we propose a novel method of using \textit{both high and low quality} BT data by providing hints (as source tags on the encoder) to the model about the quality of each source-target pair. We don't filter out low quality data but instead show that these hints enable the model to learn effectively from noisy data.
Second, we address the  problem of predicting whether a source token needs to be translated or transliterated to the target language, which is common in cross-script translation tasks (i.e., where source and target do not share the written script).
For such cases, we propose training the model with additional hints (as target tags on the decoder) that provide information about the \textit{operation} required on the source (translation or both translation and transliteration). We conduct experiments and detailed analyses on standard WMT benchmarks for three cross-script low/medium-resource language pairs: \{Hindi,Gujarati,Tamil\}$\rightarrow$English. 
Our methods compare favorably with five strong and well established baselines. We show that using these hints, both separately and together, significantly improves translation quality and leads to state-of-the-art performance in all three language pairs in corresponding bilingual settings. 
\end{abstract}

% Second, we address a problem that arises often in  low-resource MT: language pairs that do not share the written script.

%Quality of the back-translated data depends on the quality of the model used to generate it and hence can often be noisy in low-resource languages where not enough high quality bi-text data is available. Current research  either using all back translation data or  filter out low quality BT data. 

% We show that  even noisy data has partially correct information the model can learn from. 

%Methods proposed in this paper outperform SoTA and a simulated low-resource scenario of a high-resource language pair German$\rightarrow$English.

\section{Introduction}
\begin{figure}
    \centering
    \frame{\includegraphics[width=0.95\linewidth]{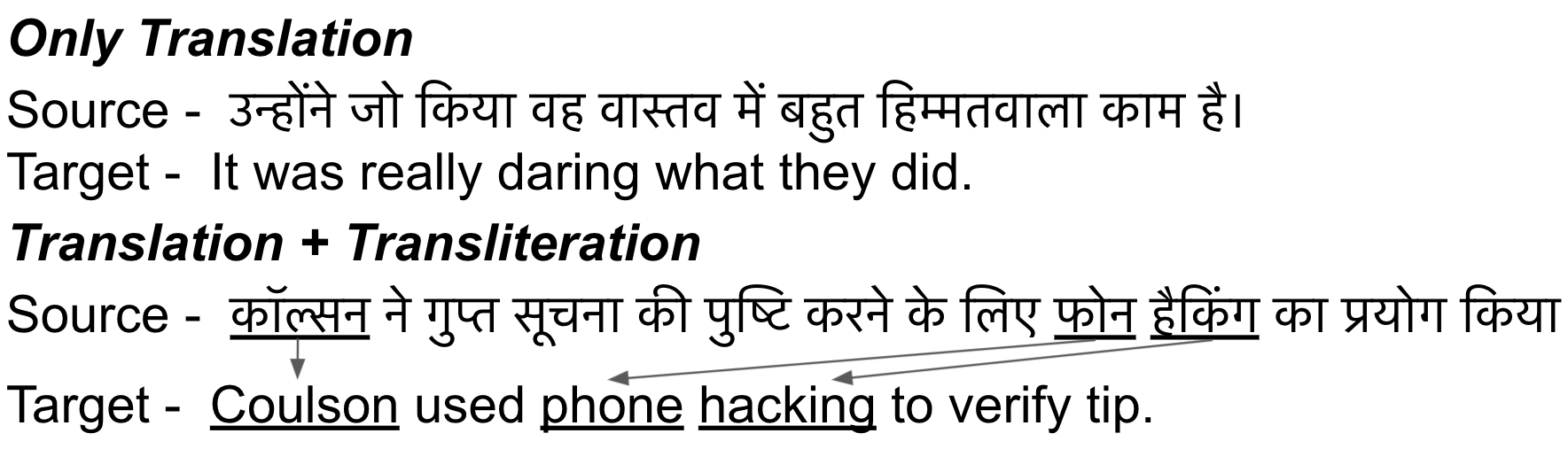}}
    \caption{Examples from the WMT 2014 Hindi$\rightarrow$English test set. The top example is a case of only translation, and the bottom one is a case where some words in the source (a named-entity "Coulson", and English words `phone', `hacking' written in Hindi) need to be transliterated.}
    \label{fig:cross_script_example}
    \vspace{-3mm}
\end{figure}
Neural machine translation (NMT) \cite{kalchbrenner-blunsom:2013:EMNLP, bahdanau2014neural, sutskever2014sequence,wu2016google, hassan2018achieving} models have become the state-of-the-art approach to machine translation. However, NMT models are data hungry and have been shown to under-perform in low-resource scenarios~\cite{koehn-knowles-2017-six}.  Various supervised and unsupervised techniques \cite{song2019mass,gulcehre2015using} have been proposed to address the paucity of high-quality parallel data in such cases. \textit{Back-translation}~\cite{sennrich-etal-2016-improving} is one such widely used data augmentation technique in which synthetic parallel data is created by translating monolingual data in the target language to the source language using a baseline system. However, in order to get high quality parallel back-translated (BT) data, we need a high quality target$\rightarrow$source translation model \cite{burlot2019using}. This in turn depends on having a substantial amount of high quality parallel (bitext) data already available. For low-resource languages, both the quantity and quality of bitext data is limited, leading to poor back-translation models.  Existing methods either use all BT data available \cite{sennrich-zhang-2019-revisiting}, or use various cleaning techniques to identify and filter out lower quality BT data \cite{khatri-bhattacharyya-2020-filtering, imankulova-etal-2017-improving}. However, filtering reduces the amount of data available for training in a scenario which is already low-resource.  How to efficiently use back-translation data in a situation where data is both scarce and of varied quality is the first key challenge we tackle in this paper. 
 
The second challenge that arises increasingly often in low-resource MT is that of cross-script NMT: translation tasks where the source and target languages do not share the same script. Cross-script NMT tasks have been steadily increasing in the WMT shared news translation tasks\footnote{http://www.statmt.org/wmt20/translation-task.html} over the past few years (28\% of tasks in 2017 and 2018, 44\% in 2019, and 63\% in 2020).  Cross-script NMT models must implicitly predict whether a source token needs to be translated or transliterated (see example in Figure \ref{fig:cross_script_example}).  Lack of shared vocabulary coupled with low data quantity and quality makes cross-script NMT in low-resource settings a very challenging task. 

\noindent In this work, we propose \textbf{HintedBT}, a family of techniques that provide hints to the model to make the limited BT data even more effective. We present results on three cross-script WMT datasets:
Hindi(hi)/Gujarati(gu)/Tamil(ta)$\rightarrow$English(en). In our first proposed HintedBT method, \textbf{\textit{Quality Tagging}}, we use tags to provide hints to the model about the quality of each source-target BT pair.  
%we show that it is possible to learn from even poor quality back-translated data. Contrary to current methods that filter out such poor quality BT data,  
%For instance, we show  a BLEU improvement of +10.6 in gu$\rightarrow$en over a baseline that uses all the back-translated data without quality hints. 
In the second method, \textbf{\textit{Translit Tagging}}, we use tags to address the cross-script NMT challenge described above: we force the decoder to predict the \textit{operation} that needs to be done on the source - only translation (or) both translation + transliteration, in addition to predicting the translated sentence.
%of being able to predict whether a source token needs to be translated or transliterated. 
% In addition to predicting the translated sentence, we force the decoder to predict the \textit{operation} that needs to be done on the source - only translation (or) both translation+transliteration. 
The correct operation is provided as an additional tag during training. 

\noindent We make the following contributions in this paper: 

%We do this by training the model with {\em transliteration hints}, i.e., relevant operation tags as the first token on the target side of the train data. 
%We see that this method provides an improvement of +6.3 BLEU in ta$\rightarrow$en over a baseline that uses all the back-translated data without transliteration hints.

%With both these motivations, we propose a novel method of providing hints to the model about the quality of the source-target pair, in the form of tags on the source sentence. With this additional information, the model can now learn from all the data available, and also learn to the appropriate extent from each data point. We call this method as \\
%We further focus on a specific subset of low resource MT, in which the source and target language do not share the same script. 
%To address this xxx, we propose training the model to predict the operation to be done on the source-  We call this method as \textbf{\textit{translit-tagging}}.

%{\color{red}{The first 1.5 points seem like a repetition of the previous two paragraphs. We can add our numerical contributions here instead.}}
\begin{enumerate}[noitemsep]
\item{Two novel hinting techniques: Quality Tagging (Section~\ref{sec:qual_tag_explain}) and Translit Tagging (Section~\ref{sec:translit_tag_explain}) to address two key challenges in low-resource cross-script MT}.
\item{Extensive experiments and comparisons to competitive baselines which show that a combination of our methods outperform bilingual state-of-the-art models for all three languages studied (Section~\ref{sec:exp_setup},~\ref{sec:analysis}). Table~\ref{tab:sota} shows BLEU scores of our methods compared to SoTA}. 
\item{Applications of proposed techniques in other situations that arise commonly in low-resource language settings (Section~\ref{sec:lr_issues}).}
%\item Second, via the {\em Quality tagging hint}, we show that it is possible to learn from poor quality  BT data as well. Also, we show via the {\em Translit Tagging Hint} that jointly predicting the translation and the cross-script operation helps improve performance of the model. We show that combining the two hints further improves the performance of our models.   Finally, we perform detailed experiments with comparisons to 4 baselines, prior work and show our model outperforms current techniques in the single language setting.
\end{enumerate}

\begin{table}[]
    \small
    \centering
    % \resizebox{0.8\linewidth}{!}{
    \begin{tabular}{|c|c|c|}
    \hline
    Lang. Pair & SoTA & This work \\ \hline
    \multirow{4}{*}{hi$\rightarrow$en} & 16.7 [Bilingual] & \multirow{4}{*}{32.0} \\
    & \cite{matthews2014cmu} & \\
    \cline{2-2}
     & 28.7 [Multilingual]   & \\
    & \cite{wang2020multi} & \\ \hline
    \multirow{4}{*}{gu$\rightarrow$en} & 18.4 [Bilingual]  & \multirow{4}{*}{20.8} \\
     & \cite{bei2019gtcom} & \\ 
    \cline{2-2}
     & 24.9 [Multilingual] & \\
     & \cite{li2019niutrans} & \\ \hline
    \multirow{4}{*}{ta$\rightarrow$en} & 15.8 [Bilingual] & \multirow{4}{*}{17.2} \\
    & \cite{parthasarathy2020adapt} & \\
    \cline{2-2}
     & 21.5 [Multilingual] & \\
     & \cite{chen2020facebook} & \\ \hline
    \end{tabular}%}
    \caption{Current SoTA versus our contributions. Our methods beat the bilingual SoTA for all three language pairs, and are competent with the multilingual SoTA, despite not using additional information in the form of pivot languages and/or multilingual models.}
    \label{tab:sota}
\end{table}

%(Need to write properly) In this work, we report results on three low resource, cross-script% language pairs - hi2en, gu2en and ta2en. Strong baselines, methods separately and in combination. Analysis with more data, iterative finetuning. The quality tagging analysis with de2en.

% We want to stress on the quality aspect \textit{and} the cross-script aspect of low resource. We can also partition the language pairs into three spaces based on the bitext and BT quality - bitext either high or low quality, and then BT either high or low quality (bitext low, BT high is probably improbable, and we don't present results for that here).

\section{Related Work}
\textbf{Leveraging monolingual data for NMT:} Initial efforts in this space focused on using target-side language models~\cite{10.5555/3015812.3015835, gulcehre2015using}. Recently, \textit{back-translation}, first introduced for phrase-based models~\cite{bertoldi-federico-2009-domain,bojar-tamchyna-2011-improving} and popularized for NMT by~\citet{sennrich-etal-2016-improving}, has been widely used. It has been shown that the quality of the back-translated data matters~\cite{hoang-etal-2018-iterative,burlot-yvon-2018-using}. Given this finding, several works have performed filtering using sentence-level similarity metrics on the round-trip translated target and the original target~\cite{imankulova-etal-2017-improving,khatri-bhattacharyya-2020-filtering}, or cross-entropy scores~\cite{junczys-dowmunt-2018-dual}. Several works have looked into iterative back-translation for supervised and unsupervised MT~\cite{hoang-etal-2018-iterative, cotterell2018explaining, niu-etal-2018-bi, lample-etal-2018-phrase, artetxe-etal-2018-unsupervised}.

\noindent \textbf{Multilingual models:} Another direction in low-data settings is to leverage parallel data from other language-pairs through pre-training or jointly training multilingual models~\cite{zoph-etal-2016-transfer, johnson2017google, nguyen-chiang-2017-transfer, gu-etal-2018-meta, kocmi-bojar-2018-trivial, aharoni-etal-2019-massively, arivazhagan2019massively}. Amongst recent WMT submissions, \citet{chen2020facebook,zhang2020niutrans,kocmi2019cuni} train multilingual models for ta$\rightarrow$en and gu$\rightarrow$en, whereas \citet{goyal2019iiit,bawden2019university,dabre2019nict,li2019niutrans} pivot through Hindi, or transliterate Hindi data to Gujarati for training gu$\rightarrow$en models. \citet{wang2020multi} train a multilingual model for hi$\rightarrow$en with a multi-task learning framework that jointly trains the model on a translation task on parallel data and two denoising tasks on monolingual data. Improving low-resource MT without leveraging data from other language-pairs has received lesser attention, 
notably in~\citet{nguyen-chiang-2018-improving, ramesh-sankaranarayanan-2018-neural, sennrich-zhang-2019-revisiting}. In this work, we experiment with bilingual models only, using no additional information from other language pairs.

\noindent \textbf{Using tags during NMT training:} Tags on the source side of NMT systems have been used to denote the target language in a multilingual system~\cite{johnson2017google}, formality or politeness~\cite{yamagishi-etal-2016-controlling,sennrich-etal-2016-controlling}, gender information~\cite{kuczmarski2018gender}, the source domain~\cite{kobus-etal-2017-domain}, translationese~\cite{riley-etal-2020-translationese}, or whether the source is a back-translation~\cite{caswell-etal-2019-tagged}. In this work, we use tags on the source side to represent the quality of the BT pair, and tags on the target side to represent the operation done on the source (translation, or translation + transliteration).
%\section{BackTranslation Quality}
%Goal is to convice reader that
%- considering quality is beneficial  (topk is better than full BT)
%- semantic is better than syntactiv  (human eval and correlation,  anecodtal examples where jaccard can fail)
%{\color{red}{TODO: experiment to correlate human eval and Jaccard / LaBSE}}
%intro: talk about quality estimation

%\subsection{Syntactic Similarity} 

%\subsection{Semantic Similarity}

%\subsection{Quality Comparison}

%\subsection {Topk vs Full BT}

 % We first 

% 1) Using all the backtranslations \\
% 2) FQE - Using topk backtranslations (as given by LaBSE or jaccard scores) - this is a stronger baseline that we are proposing.
% 3) Iterative BT
\section{Quality-based Tagging of the BT data} \label{sec:qual_tag_explain}
%The quality of generated BT data is primarily dependent on the translation model used to produce it. 
In low-resource scenarios, where bitext data is low in quantity and quality, BT data will likely contain pairs with varying quality.
%, including very low quality pairs. 
So far, there have been two broad approaches to deal with BT data: (a) full-BT: use all the BT data without considering the quality of the BT pairs ~\cite{sennrich-zhang-2019-revisiting} (b) topk-BT: use only high quality BT pairs by introducing some notion of quality between the source and target ~\cite{khatri-bhattacharyya-2020-filtering, imankulova-etal-2017-improving}. 
The full-BT method suffers from the disadvantage that it mixes the good and bad quality data, hence confusing the model. This was one of the primary motivations for introducing topk-BT models. However topk-BT models, while being {\em quality-aware}, filter away a substantial chunk of the parallel data which could be harmful in low-resource settings. 

\begin{figure}
    \centering
    \frame{\includegraphics[width=0.97\linewidth]{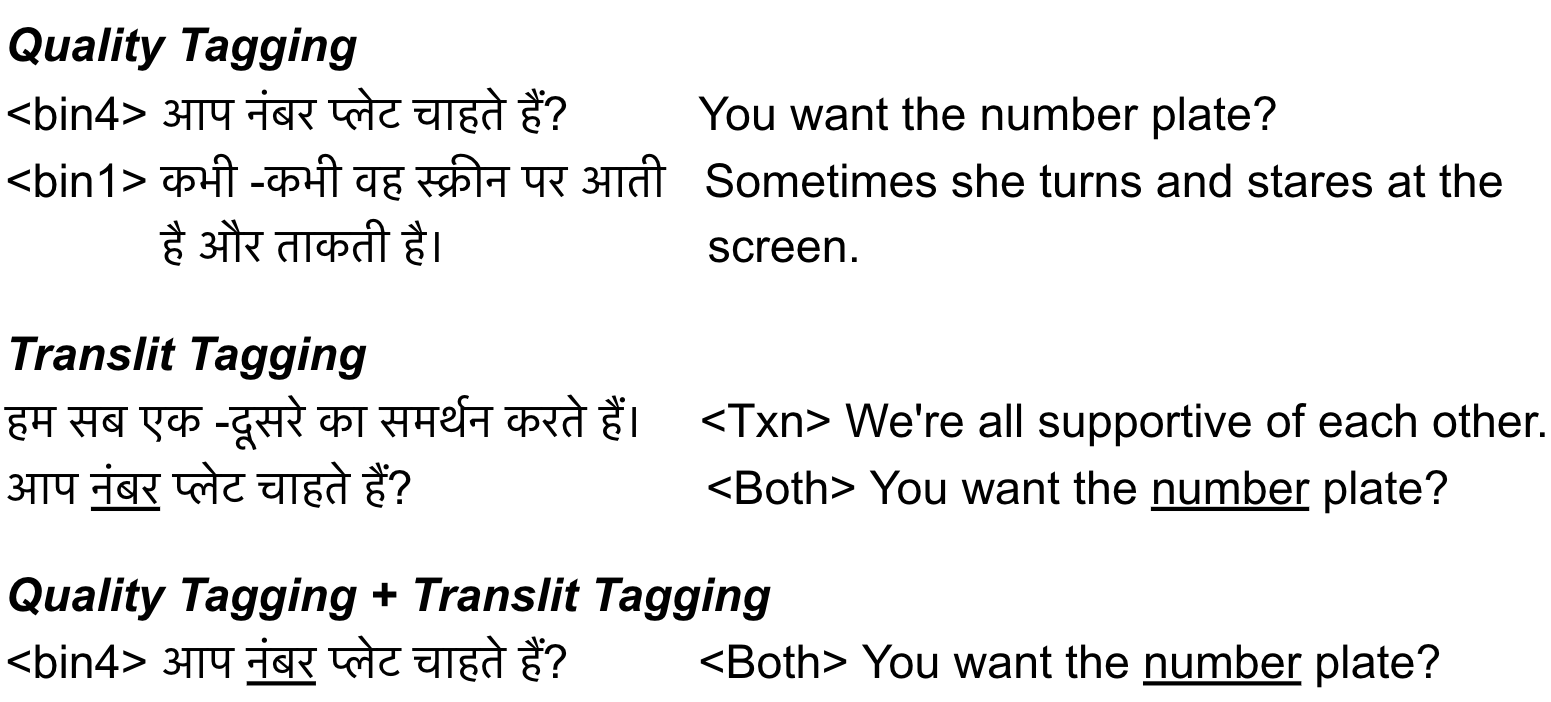}}
    \caption{Quality tags are prepended to the source, with $<$bin1$>$/$<$bin4$>$ samples being the lowest/highest quality respectively. Translit tags are prepended to the target, with $<$Txn$>$/$<$Both$>$ being  translation only or translation + transliteration respectively.
    % source sentence in the $<$bin1$>$ example translates to -
    Correct translation of $<$bin1$>$ example: \textit{Sometimes she comes on the screen and stares}.
    }
    \label{fig:tagging_demonstrate}
    \vspace{-4mm}
\end{figure}

In this work, we introduce a third type of using BT data called~\textit{\textbf{Quality Tagging}}. This approach uses {\em all} the BT data by utilizing quality information about each instance.  Our method extends the {\em Tagged BT} approach~\cite{caswell-etal-2019-tagged} that uses ``tags" or markers on the source to differentiate between bitext and BT data. We attach multiple tags to the BT data, where each tag corresponds to a {\em quality bin}. The quality bin provides a hint of the quality of the BT pair being tagged. We use  LaBSE \cite{feng2020language}, a BERT-based language-agnostic cross-lingual  model to compute sentence embeddings. The cosine similarity between these source and target embeddings is treated as the quality score of the BT pair.  BT pairs are then binned into $k$ groups based on the quality score, and the bin-id is used as a tag in the source while training (cf.~examples in Figure~\ref{fig:tagging_demonstrate}).

We explore three design choices in Quality Tagging below: a) Bin Assignment: How to assign a particular BT pair to a bin? b) Number of bins to use c) Bitext Quality Tagging.  

\noindent
\textbf{Design Choice 1 - Bin Assignment:} 
We have two direct options: {\em Equal Width Binning} or {\em Equal Volume Binning}. In {\em Equal Width Binning}, we  divide quality score range into $k$ intervals of equal size.  
%Each interval is of size $w = (max_{score} - min_{score})/k$, where $k$ is the number of bins, $max_{score}$ and $min_{score}$ are the maximum and minimum values of the similarity scores for this dataset.
Each interval then corresponds to a bin and each BT pair is assigned to the bin which contains its quality score. In {\em Equal Volume Binning}  we sort the $N$ data points by their quality score and divide points into $k$ equally sized groups. Each group then corresponds to a bin. We see that Equal Width Binning (and other size-agnostic approaches like k-means) can cause severely size-unbalanced bins, with the lowest bin(s) not adding any signal at all. This is primarily because the cosine similarity used as quality score  is language-pair agnostic and not calibrated to well separated quality bins.  Equal Volume binning addresses this concern while also providing sufficiently inherent quality-based clusters with a good choice of $k$.
%Keeping in view the size balance constraint, we did not explore  clustering methods like k-means. 

\noindent
\textbf{Design Choice 2 - Number of Bins:} 
We experimented with different number of bins (see detailed results in Appendix~\ref{app:bin_count}). From the dev-BLEU scores, we found that for hi$\rightarrow$en and gu$\rightarrow$en, four bins provide the best performance, while for ta$\rightarrow$en either three or four bins work equally well. We uniformly use four bins for the sake of simplicity and point out that deeper analysis of the interplay between bitext, BT quality and number of bins is an interesting area of future work. 
% the incremental value of adding a new bin beyond the fourth bin is minimal. So we use 4 bins with Equal Volume binning in all our experiments. 

\noindent
\textbf{Design Choice 3 - Bitext Quality Tagging:} 
%we explore whether the bitext should be quality tagged as well.
We have three choices for this question: a) Bitext is left untagged. b)Bitext is always tagged with the highest quality bin. c) A Bitext pair is also scored using LaBSE and assigned to a bin just as a BT pair would be. We discuss this design choice further in Section~\ref{sec:qual_tag_exps}. 

\begin{table*}
% \small
\centering
\resizebox{0.95\textwidth}{!}{%
\begin{tabular}{|l|c|c|c|}
\hline
    &   Train & WMT newsdev/test \\%Dev \& Test\\
   \hline
   hi$\rightarrow$en &  IIT Bombay en-hi Corpus \cite{kunchukuttan2017iit} - 1.68M pairs &   WMT-2014 (520/2.5k pairs)\\
   \hline
   gu$\rightarrow$en & WMT-2019 gu-en, TED2020 \cite{reimers2020making}, GNOME & \\
   &   \& Ubuntu \cite{tiedemann2012parallel}, OPUS \cite{zhang2020improving} - 162k pairs & WMT-2019 (3.4k/1k pairs)\\ \hline
   ta$\rightarrow$en & WMT-2020 ta-en, GNOME \cite{tiedemann2012parallel}, OPUS \cite{zhang2020improving} - 630k pairs & WMT-2020 (2k/1k pairs) \\
  % &  gu-en data from OPUS \cite{zhang2020improving} & & \\
\hline

\end{tabular}%
}
\caption{Datasets used for training. The dev and tests are used from the WMT corpus.}\label{tab:datasets}
\end{table*}

\section{Translit Tagging of the BT data} \label{sec:translit_tag_explain}
When the source and target are written in different scripts, certain words in the source explicitly need to be transliterated to the target language, such as entities, or target language words written in the source script (see example in Figure~\ref{fig:cross_script_example}).
%such as entities or target language words written in the source script as seen in the example in
In such cases, the model needs to identify which source words should be translated to the target language, and which need to be transliterated. 
To understand the prevalence of this pattern, we split the test data into two categories: {\em \{Txn, Both\}}.  {\em `Txn'} means the target sentence requires translating every source word and {\em `Both'} means the   a mix of translation and transliteration is needed to generate the target from the source words. Then we compare the percentage of sentence pairs in each category for  the hi/gu/ta$\rightarrow$en WMT test sets.  
%which require only translation versus those that require a mix of translation and transliteration. 
%We demonstrate the process we follow in Figure \ref{fig:process_of_translit}.
For each word in the source sentence, we use FST transliteration models \cite{hellsten2017transliterated} to generate 10 English (i.e., the target language) transliterations. If any of these transliterations are present in the corresponding target, we categorize the pair as {\em Both}, else as {\em Txn}. 
%mark that this source requires both translation and transliteration.
From Table~\ref{tab:mixed_stats}, we see that for all the three WMT test sets, $\sim$60-80\% of the test corpora require a mix of translation and transliteration to be performed on the source sentences. Further details about the FST models are included in Appendix \ref{app:fst_translit_models}.
%\begin{figure}
%    \centering
%    \frame{\includegraphics[width=0.97\linewidth]{images/translit4.png}}
%    \caption{We generate 10 transliterations for each source word. If any of these are present in the target, we mark that the source both translation and transliteration to the target language.}
%    \label{fig:process_of_translit}
%\end{figure}
% \vspace{-1em}

\begin{table}
\small
\centering
\begin{tabular}{|l|c|c|c|}
\hline
 \multirow{2}{*}{WMT Test Set}   & Txn: Translation &  Both: Translation + \\
  & Only & Transliteration\\
\hline
    hi$\rightarrow$en (2014) & 21.3\%& 78.7\% \\
    gu$\rightarrow$en (2019)  & 30.6\% & 69.4\% \\
    ta$\rightarrow$en (2020) & 40.5\% & 59.5\% \\
  \hline
\end{tabular}
\caption{\% of the WMT test sets where task is either only translation or a mix of translation \& transliteration}
\label{tab:mixed_stats}
% \vspace{-3mm}
\end{table}

% \vspace{-1em}
% \vspace{1mm}

To utilize this information about cross-script data in training, we propose a novel method: \textit{Translit Tagging}. We use the aforementioned methodology to split the train data into two categories: {\em \{Txn, Both\}} as before.
%{\em `Txn'} means the target sentence requires translating every source word. {\em `Both'} means the target sentence requires a mix of translation and transliteration of source words. 
We then convert this information into tags, which we prepend to the \textit{target} sentence (refer Figure \ref{fig:tagging_demonstrate} for an example). This method teaches the model to predict if the transliteration operation is required or not for the given source sentence, hence the name `translit' tagging. During inference, the model first produces the translit tag on the output, before producing the rest of the translated text. 
%We remove these tags from both the output and the reference while evaluating model performance.
Another option is to present translit tags on the source side while training.
This method does not perform as well and also has practical challenges that we describe in detail in Appendix~\ref{app:source_translit_tagging}.
%We tried an alternate setup where we predict tags indicating the translation or transliteration operation for each output word. However this did not work well due to sub word tokenization and . We present detailed experiments and results in Section \ref{sec:translit_tag_exps}. 

% Here, add the table talking about our translation versus mixed splits in the test sets. Then we talk about our method. We can just do the target decoding method, and add the source method in the appendix, or in the analysis.

\section{Experiments} \label{sec:exp_setup}

\subsection{\bf Datasets}
\noindent Table \ref{tab:datasets}~describes the train, dev, and test data used in our experiments. We  train source$\rightarrow$target and target$\rightarrow$source NMT models on the available bitext data for all language pairs. We use the latter to generate synthetic back-translation data from the WMT Newscrawl 2013 English monolingual corpus.     
%The quality of  a source-target pair  is measured as the cosine similarity between the LaBSE embeddings of the two sentences.   
 
\subsection{\bf Model Architecture} \label{subsec:Model Architecture}
We train standard Transformer encoder-decoder models as described in~\citet{vaswani2017attention}. The dimension of transformer layers, token embeddings and positional embeddings is 1024, the feedforward layer dimension is 8192, and number of attention heads is 16. We use 6 layers in both encoder and decoder for the hi$\rightarrow$en models and 4 layers for the gu$\rightarrow$en and ta$\rightarrow$en models. For training, we use the Adafactor optimizer with $\beta_1 = 0.9$ and $\beta_2 = 0.98$, and the learning rate is varied with warmup for 40,000 steps followed by decay as in~\citet{vaswani2017attention}. We perform all experiments on TPUs, and train models for 300k steps. We use a batch size of 3k across all models and tokenize the source and target using WordPiece tokenization \cite{schuster2012japanese, wu2016google}. Further details on hyper-parameter selection and experimental setup can be found in Appendix \ref{app:exp_setup_app}.
% {\color{red}{
% Add info about WPM - move here from appendix. (shared WP vocab, cite the wpm paper/BPE/GNMT paper) - 32k word pieces shared vocabulary, batch size and number of steps trained.
% }} 

\subsection{\bf Evaluation Metrics}\label{sec:metrics}
We use SacreBLEU\footnote{SacreBLEU Hash: BLEU+case.mixed+numrefs.1+\\ smooth.exp+tok.13a+version.1.5.0} \cite{post-2018-call} to evaluate our models. For human evaluation of our data, we ask raters to evaluate each source-target pair on a scale of 0-6 similar to \citet{wu2016google}, where 0 is the lowest and 6 is the highest (more details in Appendix \ref{app:human_eval}). 

\subsection{Baselines}
\begin{table*}[h]
    \small
    \centering
    \resizebox{0.97\linewidth}{!}{
    \begin{tabular}{|c|c|l|l|l|l|}
    \hline
      & $\#$ & \textbf{Modeling Methodology} & \textbf{hi$\rightarrow$en} & \textbf{gu$\rightarrow$en} & \textbf{ta$\rightarrow$en} \\ \hline
    WMT Data & 1 & bitext & 19.5 & 8.4 & 11.3 \\
    % \cline{2-6}
    Baselines & 2 & bitext + full-BT & 30.9 & 15.0 & 14.1\\ \Xhline{4\arrayrulewidth}
    \multirow{2}{*}{Prior} & 3 & bitext + Iter-BT & 29.2 & 16.5 & 14.9 \\ 
    %  \cline{2-6}
    \multirow{2}{*}{Work} & 4 & bitext + tagged-full-BT & 30.2 & 17.0 & 16.0 \\
    % \cline{2-6}
     & 5 & bitext + \textit{LaBSE} topk-BT & 31.2 & 16.0 & 16.4 \\ \Xhline{4\arrayrulewidth}
     \multirow{3}{*}{HintedBT} & 6 & bitext + full-BT + \textit{LaBSE} quality tags & 31.2 & 17.6 & 15.5 \\
     & 7 & bitext + full-BT + translit-tags & 31.0 & 15.2 & 15.0 \\
 %   8 & bitext + \textit{LaBSE} topk-BT + translit-tags & 30.9  & 16.3 & 16.2 \\ \hline
     & 8 & bitext + full-BT + \textit{LaBSE} quality tags + translit-tags & \textbf{31.6} & 17.9 & 16.0 \\ \Xhline{4\arrayrulewidth}
     \multirow{2}{*}{Iterative} & 9 & bitext + tagged-Iter-BT & 30.0 & 20.5 & 16.5 \\
     \multirow{2}{*}{HintedBT} & 10 & bitext + Iter-BT + \textit{LaBSE} quality tags & 29.9 & 20.0 & \textbf{17.2} \\
      & 11 & bitext + Iter-BT + \textit{LaBSE} quality tags + translit-tags & 29.5 & \textbf{20.8} & 16.3 \\ \hline
    \end{tabular}}
    \caption{Performance of models on WMT test sets.}
    \label{tab:main_results}
\end{table*}

% Rows 1-5 are baseline models, row 5 depicts the quality tagged model, rows 6-7 depict translit tagged models, and row 8 shows the combined quality+translit tagged model.

We present the following five baseline models to compare our methods against. Baselines 3-5 are our re-implementations of relevant prior work which introduce different methods of improving on the full-BT baseline (Baseline 2).
%Baselines 1 and 2 are based on data available in WMT. Baselines 3-5 are our reimplementation of relevant prior work. 
\begin{enumerate}[noitemsep]
    
   \item  \textbf{bitext} - Model trained only on bitext data. The size of train data is shown in Table~\ref{sec:exp_setup}. 
    %There are 1.6M such pairs in hi$\rightarrow$en, 168k in gu$\rightarrow$en and 630k in 630k in ta$\rightarrow$en.
     \item \textbf{bitext + full-BT} - Model trained on bitext data and an additional 23M back-translated pairs.
     %generated from the WMT Newscrawl 2013's English monolingual data.
    \item \textbf{bitext + Iterative-BT} - Iterative training of models in the forward and reverse directions~\cite{hoang-etal-2018-iterative}. In our experiments, models are trained with two iterations of back-translation. 
   We also study the interaction of Iterative-BT with HintedBT in Section \ref{sec:iterative_bt}.   
    \item \textbf{bitext + tagged-full-BT} - Model trained on bitext data and tagged full-BT data \cite{caswell2019tagged}.  A tag is added to the source in every BT pair to help the model distinguish between natural (bitext) and synthetic (BT) data. 
    \item \textbf{bitext + LaBSE topk-BT} - Model trained on bitext data and {\em topk} best quality BT pairs. Quality is estimated using LaBSE scores, and we grid-search with at least 6 LaBSE threshold values and choose the one which gives the best BLEU on the dev set (see Appendix \ref{sec:appendix_baseliens} for more details). The chosen threshold yields 20M BT sentences for hi$\rightarrow$en, 10M for gu$\rightarrow$en and 5M for ta$\rightarrow$en. 
    %We use the computed LaBSE-based similarity scores to filter out the lower quality BT pairs. 
    %For each language pair, we individually pick the best top BT data through  experimentation with different LaBSE score thresholds (refer). 
 %Note that to get the best topk model, we run 6-7 experiments for each language pair, making this a strong (and computationally expensive) baseline.
\end{enumerate}

\begin{table}
    \small
    \centering
    \begin{tabular}{|l|l|l|l|}
    \hline
    \textbf{Data Quality} & \textbf{hi$\rightarrow$en} & \textbf{gu$\rightarrow$en} & \textbf{ta$\rightarrow$en} \\ \hline
    bitext & 4.16$\pm$0.15 & 4.37$\pm$0.15 & 3.73$\pm$0.17 \\
    full-BT & 4.41$\pm$0.11 & 3.42$\pm$0.14 & 3.51$\pm$0.13 \\ \hline
    \end{tabular}
    \caption{Mean human quality scores on 500 samples of bitext and full-BT data alone with 95\% CI. Integer ratings for individual sentence pairs lie in [0,6].}
    \label{tab:qual_scores}
\end{table}

\noindent We report the performance of these baseline models on the WMT test sets in rows 1-5 in Table \ref{tab:main_results}.
Adding BT data alone (Row-2) provides a significant improvement in performance for  hi$\rightarrow$en ($+58\%$) and  gu$\rightarrow$en ($+78\% $) over the plain bitext baseline (Row-1).  However for  ta$\rightarrow$en, the improvement is comparatively smaller ($+24.7\%$). To understand this deviation further, we conduct a human evaluation (Section~\ref{sec:metrics}) on a random 500 samples of the bitext data. The results are reported in Table~\ref{tab:qual_scores}. We see that the ta$\rightarrow$en bitext data is much poorer in quality compared to the other two pairs. This affects the quality of the back-translation model and hence influences the results of a few more experiments we report further. 
Next we see that iterative back-translation (Row-3) and tagged back-translation (Row-4) do improve the performance for gu$\rightarrow$en, ta$\rightarrow$en but not for hi$\rightarrow$en, when compared to Row-2.  Comparison between full-BT (Row-2) and topk-BT (Row-5) shows 
%that BT quality does play a role in the final quality of the model. 
choosing high quality BT data instead of using all the BT data proves beneficial for all 3 language pairs.

\begin{table}[]
    \centering
    \resizebox{0.97\linewidth}{!}{
    \begin{tabular}{|l|l|l|l|l|}
    \hline
         & \textbf{Bin 1} & \textbf{Bin 2} & \textbf{Bin 3} & \textbf{Bin 4}\\ \hline 
        hi$\rightarrow$en & 4.05 $\pm$0.13 & 4.47 $\pm$0.10 & 4.53 $\pm$0.10 & 4.87 $\pm$0.09  \\
        gu$\rightarrow$en & 2.28 $\pm$0.17 & 2.79 $\pm$0.15 & 3.18 $\pm$0.15 & 3.85 $\pm$0.15 \\
        ta$\rightarrow$en & 1.44 $\pm$0.17 & 2.78 $\pm$0.15 & 3.31 $\pm$0.14 & 3.99 $\pm$0.13 \\
        \hline
    \end{tabular}}
    \caption{Mean human quality scores for the 4 quality bins along with 95\% CIs. %Integer ratings for individual sentence pairs lie in [0,6].
    }
    \label{tab:bins_human}
\end{table}

\subsection{Quality Tagging} \label{sec:qual_tag_exps}
As explained in Section \ref{sec:qual_tag_explain}, we assign each BT pair to one of four quality bins that have equal volume of pairs in them. Table~\ref{tab:bins_human} presents the mean quality score as annotated by humans for different bins. We see a perceptible difference in the quality of data across bins for all languages. This confirms our hypothesis that BT data will be of varied quality. It reinforces faith in our choice of four equal volume bins and also in LaBSE as a method for automatic quality evaluation.

We now explore the choice of how to tag the bitext data (i.e., design choice 3), using human evaluation of the bitext and BT data (see Table \ref{tab:qual_bin_base}). To re-iterate, we perform human evaluation of data by having raters evaluate each source-target pair on a scale of 0-6, with 0 being the lowest, and 6 being the highest (more details in Appendix \ref{app:human_eval}). For hi$\rightarrow$en, both bitext and BT data are of high quality (>4). Hence, we decide to tag the bitext with the highest quality bin $<$bin4$>$. For gu$\rightarrow$en, the BT data is of lower quality compared to the bitext. Hence, we decide to leave the bitext untagged, making the BT data's quality tags both an indicator of quality, as well as an indicator that the data is synthetic. For ta$\rightarrow$en, both bitext and BT are of lower quality (<4), with the bitext's quality being slightly higher. Hence, here as well, we decide to leave the bitext untagged. We further demonstrate our choices using experiments. From Table~\ref{tab:qual_bin_base}, we see that for hi$\rightarrow$en, tagging with $<$bin4$>$ works best, while for gu$\rightarrow$en leaving it untagged works best. For ta$\rightarrow$en, there is no clear winner.
%melvin edit below

% These results are further validated by the human evaluation scores in Table \ref{tab:qual_scores}. For hi$\rightarrow$en, both bitext and BT data are of high quality (>4). This explains why assigning $<$bin4$>$ to the bitext works best.  For gu$\rightarrow$en, the BT data is of lower quality compared to the bitext and hence leaving it untagged works best.  
%The quality tags indicate the quality, and also differentiate between bitext and BT data leading to the observed improvements.
% For ta$\rightarrow$en, both bitext and BT are of lower quality (<4). With no clear trends from both human evaluation and empirical analyses, we decide to leave the bitext untagged.
% , since distinguishing between bitext and BT using tags helped ta$\rightarrow$en in tagged-BT (Row-4).
% For ta$\rightarrow$en, the bitext and the BT data are both of a lower quality; we keep the bitext data untagged in all our experiments. 
For the remaining experiments in this paper, we stick to this assignment for bitext tagging: gu$\rightarrow$en and ta$\rightarrow$en (untagged), hi$\rightarrow$en ($<$bin4$>$ tag).

\begin{table}[t]
    \small
    \centering
    \begin{tabular}{|l|l|l|l|}
    \hline
    \textbf{Bitext tagging} & \textbf{hi$\rightarrow$en} & \textbf{gu$\rightarrow$en} & \textbf{ta$\rightarrow$en} \\ \hline
     Untagged & 30.0 & 17.6 & 15.5 \\
     Tagged with $<$bin4$>$& 31.2 & 16.8 & 15.6 \\ 
     LaBSE Quality Tags & 30.9 & 16.2 & 15.7 \\ \hline
    \end{tabular}
    \caption{Quality tagging on full-BT data, with bitext tagged/untagged}
    \vspace{-5mm}
    \label{tab:qual_bin_base}
\end{table}

%- we analyze this further in Section \ref{yet to add}. These decisions are further corroborated by the performance of the baseline tagged BT models -hi$\rightarrow$en requires bitext and BT to be treated the same, and gu/ta$\rightarrow$en require the BT to be marked as synthetic.

We present results of the quality tagged models in Row-6 of Table ~\ref{tab:main_results}. First when we compare Row-6 with full-BT in Row-2, we see that quality tagging always yields higher BLEU. Same pattern exists with Row-3 where quality tagging always outperforms Iterative-BT for all language pairs. This is an important result because, while Iterative-BT is effective, it is also very computationally expensive. Quality tagging is able to produce better results than Iterative-BT with far lesser computational costs.  Quality Tagging again outperforms both tagged-BT and topk-BT for hi$\rightarrow$en and gu$\rightarrow$en. For ta$\rightarrow$en, topk-BT still has the best BLEU. We delve into more details on why this happens in Section~\ref{sec:ta2en_further_analysis}.  To summarize, we see quality tagging provides the best performance across all previous baselines (except in two ta$\rightarrow$en instances). In addition, quality tagging is far more efficient than topk-BT in terms of computational resources since topk-BT requires multiple models to be trained for the threshold parameter search. 

\subsection{Translit Tagging} \label{sec:translit_tag_exps}
As explained in Section \ref{sec:translit_tag_explain}, we train the decoder to generate the translit tag (\textit{`Txn'} or \textit{`Both'}) along with the target sentence.  
% The translit tag denotes the operation performed on the source to derive the target sentence and is one of {\em \{Txn, Both\}}. 
% {\em Txn} means the target sentence is obtained by translating every source word. {\em Both} means the target sentence is obtained by a mix of doing both translation and transliteration of source words. 
%predict whether or not transliteration is required to translate the source, by attaching the corresponding translit tag to the target while training. 
During evaluation, we remove the translit tag which the model has produced in the output. 
%We tag both the bitext and the BT, since this method teaches the model to identify the operation to be done on the source. 
Row-7 in Table \ref{tab:main_results} shows the BLEU of the translit tagging models, and the corresponding baseline is Row-2. As we can see, translit tagging improves the performance of all three language-pairs over the baseline.  
%We perform two experiments with this method - one where we tag the full-BT data, and one where we tag the topk-BT data (i.e., adding translit tags to rows 2 and 4 respectively in Table \ref{tab:main_results}). We report the performances of the translit-tagged models as experiments 6 and 7 in Table \ref{tab:main_results}. 
%We compare row 6 with row 2, and row 7 with row 4 (i.e., each experiment with its corresponding baseline). For gu$\rightarrow$en, adding translit tags improves upon the full-BT baseline by +0.2 and the topk-BT baseline by +0.3 BLEU points. For ta$\rightarrow$en, adding translit tags improves upon the full-BT baseline by +0.9, but underperforms the topk-BT baseline by -0.2 BLEU points. For hi$\rightarrow$en, full-BT with and without translit tags are almost equivalent, and topk-BT with translit tags is 0.3 BLEU points lesser than the topk-BT baseline. We explore this method further in the next Section.

\subsection{HintedBT: Quality + Translit Tagging} \label{sec:qual_and_translit_exps}
We combine our methods of \textit{Quality Tagging} and \textit{Translit Tagging} in this experiment: we tag the source with quality tags (as per Section \ref{sec:qual_tag_exps}), and we tag the target with translit-tags (as per Section~\ref{sec:translit_tag_exps}).
%Again, we remove the translit-tags from the models' output and the reference during inference. 
We report the results as Row-8 in Table \ref{tab:main_results}. We see that for all 3 language pairs, the combination of these 2 methods outperforms both methods individually (comparing with Rows 6 and 7). For hi$\rightarrow$en, this combination gives the overall best results of \textbf{31.6}, and to the best of our knowledge, this outperforms the bilingual SoTA \cite{matthews2014cmu} as well as the multilingual SoTA \cite{wang2020multi}  for hi$\rightarrow$en. 
For gu$\rightarrow$en, the combination produces +1.9 over an already strong topk-BT baseline. 
However for ta$\rightarrow$en,  topk-BT still remains as the best method thus far. %even though it beats the quality-tagging and translit-tagging methods applied individually. 
%We investigate this observation further in Section \ref{sec:ta2en_further_analysis}.
%We analyze the results for ta$\rightarrow$en separately in 
% We propose a combination, that obtains better results...
% Source side quality tags, and target side translit tags

\subsection{Iterative HintedBT} \label{sec:iterative_bt}

%\begin{table}[h]
%    \centering
%    \resizebox{0.95\linewidth}{!}{
%    \begin{tabular}{|c|l|l|l|l|l|}
    %\hline
    %$\#$ & \textbf{Data} & \textbf{hi$\rightarrow$en} & \textbf{gu$\rightarrow$en} & \textbf{ta$\rightarrow$en} \\ \hline 
    %1 & bitext & 19.5 & 8.4 & 11.3 \\
    %2 & bitext + full-BT & 30.9 & 15.0 & 14.1\\ 
    %3 & bitext + Iter-BT & 29.2 & 16.5 & 14.9 \\ \hline 
     %\multirow{2}{*}{-} \\    
    % & \textbf{quality tags} &  &  &  \\ \hline
    %4 & bitext + full-BT & 31.2 & 17.6 & 15.5 \\
    %5 & bitext + Iter-BT & 29.9 & 20.0 & \textbf{17.2} \\ \hline 
    % & \textbf{With quality \&} & \multirow{2}{*}{-} & \multirow{2}{*}{-} & %\multirow{2}{*}{-} \\
  %   & \textbf{translit tags} & & & \\ \hline
  %  6 & bitext + full-BT & \textbf{31.6} & 17.9 & 16 \\
  %  7 & bitext + Iter-BT & 29.5 & \textbf{20.8} & 16.3 \\ \hline
  %  \end{tabular}}
 %   \caption{Experiments with Iterative Back-Translation.}
%    \label{tab:exps_iterative_finetuning}
%\end{table}

%In Section \ref{sec:exp_setup}, we applied our proposed methods on vanilla back-translation data (i.e., the full-BT data). Vanilla BT is widely used, but it has certain disadvantages for low-resource scenarios - since the available bitext data is low, the reverse model used to generate the BT is also of low quality, which further means that the generated BT itself is of a lower quality. 
In this section, we apply Iterative Back-Translation  \cite{hoang-etal-2018-iterative} in combination with the two methods in HintedBT: Quality Tagging and Translit Tagging. The goal here is to understand if our method is able to capitalize on the gains of Iterative-BT or whether its gains are subsumed by a powerful method like Iterative-BT. We run Iterative-BT first with quality tagging alone, and next Iterative-BT with the combination of both quality tagging and translit tagging. As an additional baseline, we also run Iterative-BT with back-translation tagging as in  Row-4 \cite{caswell2019tagged}.
We run two iterations of back-translation in all experiments.  
%is a very widely used NMT training methodology, wherein multiple rounds of BackTranslation and Forward Translation are applied, each round improving upon the previous. 
%While Iterative BT has proven to be effective for improving quality,  it computationally expensive  since it involves multiple rounds of training models in both directions.
%We run experiments with Iterative-BT, both individually and in combination with our methods. 

\noindent We perform quality tagging for models in both directions using the Equal Volume method with four bins. In every round, we generate the BT, compute the LaBSE scores and assign each pair to the right bin and train the model. 
Row-10  in Table~\ref{tab:main_results} shows BLEU when quality tagging is applied along with Iterative-BT. Comparing Row-10 with its corresponding full-BT  baseline in Row-6,  we see that the iterative version performs even better, with gu$\rightarrow$en and ta$\rightarrow$en getting BLEU scores of \textbf{20.0} and \textbf{17.2}, respectively. To the best of our knowledge, this outperforms the bilingual SoTA for ta$\rightarrow$en \cite{parthasarathy2020adapt}.

% While training the translit tagging addition to the above model, we note that the translit tag is the same irrespective of the direction of translation. 
Row-11 shows the performance when Iterative-BT is combined with both Quality Tagging and Translit Tagging. Comparing Row-11 with its corresponding full-BT baseline in Row-8, we see that this helps for gu$\rightarrow$en, giving a further boost in performance of +0.8 to get a final BLEU score of \textbf{20.8}. To the best of our knowledge this outperforms the bilingual SoTA performance for gu$\rightarrow$en \cite{bei2019gtcom}. 
%\textbf{(Rows 2, 3, 4)} We see that Iterative-BT is better than vanilla BT for gu/ta$\rightarrow$en as expected; however it is worse than vanilla BT for hi$\rightarrow$en. Further, we see that our method of quality tagging on vanilla BT by itself is better than Iterative-BT for all 3 language pairs (row 4 vs 3); which means that our proposed method gets better results with lower computational requirements.
To summarize, except for hi$\rightarrow$en,    Iterative-BT helps improve Hinted BT significantly. For  hi$\rightarrow$en, even plain Iterative-BT does not help as seen in Row-3. Further investigating the cause of this result is delegated to future work.  
%As future work we aim to study theinteraction between the dataset  and our experiment setting to understand why this trend does not hold.  

\section{Experiment Analysis}\label{sec:analysis}
In this section, we analyse a few key aspects of the experiments described in the previous section. 
 
 \subsection{Uniqueness of ta$\rightarrow$en} \label{sec:ta2en_further_analysis}
We observed in Section ~\ref{sec:qual_tag_exps} that  Quality Tagging  does not surpass the performance of the topk filtering strategy only for ta$\rightarrow$en. In this section we investigate this observation further. Ta$\rightarrow$en has two significant differences compared to the other two language pairs. First, from Table~\ref{tab:qual_scores} we see that the bitext quality of  ta$\rightarrow$en is much poorer. Second, only 22\% of the 23M BT data is present in topk-BT for ta$\rightarrow$en, compared to 87\% and 43\% for hi$\rightarrow$en and gu$\rightarrow$en respectively.
We posit that the large fraction of poor quality BT data interferes with the model learning from the bitext and high quality filtered BT data used in the topk-BT setting. In order to study this hypothesis, we train a model on a combination of 3 datasets: 630K of bitext, the 5M topk-BT, and 10M pairs randomly selected from the remaining 18M BT data. In total, we have 15M BT  and 630K bitext pairs. To be consistent, we perform quality binning as in Section~\ref{sec:qual_tag_exps}. In this setting, the model gets a BLEU score of \textbf{16.6}, outperforming the topk-BT method by $+0.2$ BLEU points. We repeat the above experiment by sampling 12 M noisy BT data (instead of 10M in the above set up). This drops the BLEU by $0.3$ points.     

 Hence we see that the overarching trends of being able to learn from poor quality data via quality tagging also holds for ta$\rightarrow$en. However   ratio between good and poor quality BT data is important to achieve this improvement; especially when the bitext data is of poor quality. Understanding this interaction in more depth is left to future work.

\subsection{Randomized Bin Assignment} 
In order to study the efficacy of Quality Tagging, we perform an experiment where instead of using {\em Equal Volume Binning} to choose bins (Row-6 of Table~\ref{tab:main_results}), we randomly  assign every BT pair to one of four bins. 
% Since Quality Binning has significant  impact in  gu$\rightarrow$en and ta$\rightarrow$en  (Row-6 of Table~\ref{tab:main_results}), we perform this experiment on these two language pairs. 
We observe that BLEU of hi$\rightarrow$en, gu$\rightarrow$en and ta$\rightarrow$en drops to 30.6, 16.8 and 15.9 respectively. In summary, we see that random bin assignment degrades performance of Quality Binning to almost match that of Tagged-BT.   

\subsection{Prediction of Translit Tags}
%In Sections \ref{sec:translit_tag_exps} and \ref{sec:qual_and_translit_exps}, we remove the translit tags from the model's output and the reference during evaluation.  
As mentioned in Section~\ref{sec:translit_tag_explain}, one of the key problems in cross-script NMT is to know when to translate, or transliterate a source word. In this section, we study the performance of our techniques in solving this problem. % First, we treat the task of predicting the right operation aka the TranslitTag as a classification problem and report the F1 score of a model.   Specifically we treat the tag  {\em Txn} as the positive label and {\em Both} as the negative label.  Second, 
We pose the decision of translate vs transliterate as a binary classification problem as follows: comparing the source and target, we assign a binary label to every word in the source - {\em true} if it needs to be translated, {\em false} if it needs to be transliterated. Every NMT model we train is seen as a classifier that decides whether to translate/transliterate a word; we measure its F1 score that we call as `word-level F1' (reported in Table~\ref{tab:translit_f1_score}).  

%In this section, we measure the accuracy of the translit tags predicted by the model. We report the \% accuracies in Table \ref{tab:tag_pred_accuracy}.  We see the trend that a model that is capable of predicting translit tags more accurately also has better BLEU performance in Table~\ref{tab:main_results}. {\color{red}{ The Pearsons correlation coefficient between }} 

\begin{table}[t]
    \centering
    \resizebox{0.95\linewidth}{!}{
    \begin{tabular}{|c|l|l|l|l|}
    \hline
    $\#$ & \textbf{Data} & \textbf{hi$\rightarrow$en} & \textbf{gu$\rightarrow$en} & \textbf{ta$\rightarrow$en} \\ \hline
    1 & bitext + full-BT & 77.3 & 62.2 & 56.9 \\ \hline
    2 & bitext + full-BT + translit-tags & 77.8 & 62.2 & 58.9 \\ \hline
    \multirow{2}{*}{3} & bitext + full-BT + LaBSE  & \multirow{2}{*}{78.0} & \multirow{2}{*}{66.0} & \multirow{2}{*}{59.8} \\ 
     & quality tags + translit-tags & & & \\ \Xhline{4\arrayrulewidth}
    4 & Correlation with BLEU & 0.81 & 0.99 & 0.97 \\ \hline
    \end{tabular}}
    \caption{\%Word-level F1 scores of models in transliterating  correct source words accurately. Row-4 shows Pearson's correlation of F1 scores with corresponding  BLEU scores in Table \ref{tab:main_results}.}
    \label{tab:translit_f1_score}
\end{table}

% The first 3 rows of Table~\ref{tab:translit_f1_score} show the word-level F1 score for different models. 
\noindent In Table~\ref{tab:translit_f1_score}, we see that models based on Translit Tags (Row-2)  and Quality Binning + Translit Tags (Row-3)  have equal/better F1 scores than the full-BT model (Row-1) across all languages. We  also observe that adding quality tags, though unrelated to transliteration,  helps improve the word-level F1. We compute Pearson correlation between the word-level F1 scores with corresponding model BLEU scores (Row-4). As seen, there is a very strong correlation between these two variables, confirming that adding quality tags leads to better translate vs transliterate decisions. The combination of these two factors partially explain why the two hints lead to additive BLEU gains seen in Row-8 of  Table~\ref{tab:main_results}.

\subsection{Meta-Evaluation of Results} \label{sec:meta_eval}
In this section, we perform meta-evaluation of our results using human evaluation and statistical significance tests as suggested by the guidelines in \citet{marie2021scientific}. We compare each language pair's best non-iterative model (test system) and topk-BT model (base system) in Table \ref{tab:meta_evaluation}. We first report their BLEU scores computed using SacreBLEU. 

Then, we compute human evaluation scores for both the base and test systems (using the same metric described in Section \ref{sec:metrics}). We have three human raters compare the base and test system translations using 500 randomly chosen source sentences from the test set. We report the difference in scores between the two systems (the Side-by-Side, i.e., SxS score) as the human evaluation metric. A SxS score of $\pm 0.1$ between the two systems is considered significant. We see in Table \ref{tab:meta_evaluation} that hi$\rightarrow$en and gu$\rightarrow$en have sufficient SxS scores, whereas ta$\rightarrow$en falls a little short of $0.1$.

Finally, we perform statistical significance tests to compare the base and test systems (as described in \citet{koehn2004statistical}). We create 1000 test sets with 500 random test datapoints each and calculate the two models’ SacreBLEU scores. We use the resultant SacreBLEU scores to conduct T-tests\footnote{scipy.stats.ttest\_ind}. For all three language pairs, we see significant T-statistics (reported in Table \ref{tab:meta_evaluation}) which have p-values $<0.001$.

\begin{table}[]
    \centering
    \resizebox{0.95\linewidth}{!}{
    \begin{tabular}{|l|l|l|l|}
    \hline
    \textbf{Metric} & \textbf{hi$\rightarrow$en} & \textbf{gu$\rightarrow$en} & \textbf{ta$\rightarrow$en} \\ \hline
    Best (non-iterative) & \multirow{2}{*}{31.6} & \multirow{2}{*}{17.9} & \multirow{2}{*}{16.6} \\ 
    model's BLEU & & & \\ \hline
    topk-BT BLEU & 31.2 & 16.0 & 16.4 \\ \hline
    Side-by-Side & \multirow{2}{*}{0.11} & \multirow{2}{*}{0.19} & \multirow{2}{*}{0.05} \\
    Human Eval. & & & \\ \hline
    T-statistic & 11.05 & 64.03 & 8.43 \\
    \hline
    \end{tabular}}
    \caption{Meta-Evaluation of results. In this table, we compare each language pair's best non-iterative model (test system) and topk-BT model (base system) using three metrics - BLEU scores, SxS human evaluation scores, and T-statistics from statistical significance tests.}
    \label{tab:meta_evaluation}
\end{table}

\section{Issues in Low-Resource Settings}\label{sec:lr_issues}
In this section, we discuss three issues that arise in low-resource settings, that are relevant to HintedBT. A language can be low resource if it (a) does not have enough bitext data (Section~\ref{sec:bitext_simulation}) or (b) is not well represented in open multilingual word / sentence embedding models (Section~\ref{sec:bot_jaccard}). Further, in scarce bitext settings, does having a large monolingual target corpus help HintedBT? (Section~\ref{sec:more_bt}). 
%We study the interaction of these three issues with HintedBT in this section. 
\subsection{Low Bitext Quantity Simulation}\label{sec:bitext_simulation}
\begin{table}[t]
    \centering
    \resizebox{0.97\linewidth}{!}{
    \begin{tabular}{|l|l|l|l|l|}
    \hline
        \multirow{2}{*}{\textbf{Data Used} $\downarrow$} & \multicolumn{4}{|c|}{\textbf{Bitext data size} $\rightarrow$}\\
        \cline{2-5}
         & \textbf{500k} & \textbf{200k} & \textbf{100k} & \textbf{50k} \\ \hline 
        bitext & 28.6 & 24.5 & 18.1 & 0.5 \\ \hline
        bitext + full-BT & 33.7 & 31.2 & 27.4 & 1.5  \\ \hline
        bitext + full-BT +  & 36.6 & 34.3 & 30.9 & 3.1 \\
        \cline{2-5} 
        LaBSE qual. tags & (+8.6\%) & (+9.9\%) & (+12.8\%) & (+106.6\%) \\ \hline
    \end{tabular}}
    \caption{Quality Tagging on simulated low resource scenarios of de$\rightarrow$en}
    \label{tab:german_exps}
\end{table}
Inspired by the experimentation methodology in~\citet{sennrich-zhang-2019-revisiting}, we simulate different levels of low resource conditions using a high resource language pair German(de)$\rightarrow$English(en). From the 38M bitext data points in  de$\rightarrow$en WMT 2019 news translation task, we randomly choose 500K, 200K, 100K, 50K bitext data points to simulate different low-resource scenarios. From 23M sentences of WMT 2013 Newscrawl's English monolingual data, we generate BT data and benchmark both full-BT and quality tagging on it. BT data is generated with en$\rightarrow$de models trained on the restricted bitext for each setting. We use all of the 23M BT pairs since English monolingual data is easily available and we wanted to keep the setup as realistic as possible.   Results  in  Table \ref{tab:german_exps} clearly show that quality binning outperforms full BT  under all scenarios. More interestingly, the effectiveness of quality tagging increases as the low-resourcedness increases. This shows quality tagging is able to use all the data as full-BT, but more effectively, a very desirable characteristic in a low-resource setting. 

%For quality tagging, we perform another experiment with simulated low-resource conditions of a high-resource language pair: German(de)$\rightarrow$English(en) (which shares the same script). We pick de$\rightarrow$en bitext data from- there are 38M bitext data points, as opposed to 1.68M/162k/630k bitext data points in hi/gu/ta$\rightarrow$en respectively. We simulate four low resource scenarios, choosing 50k, 100k, 200k and 500k bitext data points respectively. 
%We use  (23M sentences) to generate back-translations from. Further, we apply quality tagging (equal volume binning of LaBSE similarity scores with 4 bins, bitext untagged) to the BT data. We report results in
%We see as the situation becomes more low resource (i.e., as bitext data size decreases), the \% improvement given by quality tagging over the full-BT baseline increases.

\subsection{Quality Metric for Extremely Low Resource Languages}\label{sec:bot_jaccard}
LaBSE scoring \cite{feng2020language} depends upon the availability of the pre-trained embedding model. Some very low-resource languages may not have multilingual embeddings or, even if present, may not have high quality embeddings. 
% Quality tagging cannot then be applied to these languages. 
One alternative is to use round-trip-translation ~\cite{khatri-bhattacharyya-2020-filtering} and a syntactic comparison between the original target and the round-trip target. We use  the Jaccard similarity index \cite{huang2008similarity} between character tri-gram sets as the syntactic similarity measure. We call this measure the Bag of Trigram Jaccard or BoT-Jaccard in short.  

We study BoT-Jaccard vs LaBSE in more detail in Appendix~\ref{app:jacc_labse_qual_metric}. We summarize the results as follows. BoT-Jaccard has weaker correlation to human judgement of similarity compared to LaBSE. In our study of the failure patterns, most failures stem from the syntactic nature of the metric.  Despite the above drawbacks of BoT-Jaccard over LaBSE, we see that it performs almost on par with LaBSE and hence is a very good alternative when LaBSE is not available. We repeat all our experiments with   BoT-Jaccard and we see following improvements on the full-BT baseline. We get BLEU increases of $0.4$ for hi$\rightarrow$en, $2.8$ for gu$\rightarrow$en using quality tagging, and $1.4$ for ta$\rightarrow$en using topk-BT. 
% though in general it performs poorer than when using LaBSE, the drop is not drastic. We obtain a BLEU of 31.3 on hi$\rightarrow$en and 17.8 on gu$\rightarrow$en with quality tagging of full-BT, and a BLEU of 16.5 on ta$\rightarrow$en with topk-BT. 
% In Table~\ref{tab:metric_corr} in Appendix ~\ref{app:jacc_labse_qual_metric} we also compare correlation of LaBSE and BoT-Jaccard scores against human judgement of quality for BT data.  

\subsection{Does a larger monolingual corpus help? }\label{sec:more_bt}
In this section, we analyze if providing more BT data helps the model. 
% We run all our back-translation experiments with WMT 2013 Newscrawl's monolingual English data, which has 23M English sentences. 
We re-run  HintedBT experiments from Section \ref{sec:exp_setup} with monolingual data from both Newscrawl 2013 and 2014, resulting in a total of 46M BT pairs. We report results in Table \ref{tab:exps_more_data}. For hi$\rightarrow$en, quality tagging improves BLEU to \textbf{32.0} (an increase of 0.4 from our previous best of 31.6). For gu$\rightarrow$en and ta$\rightarrow$en, quality + translit tagging delivers performances of 18.2 and 16.1, +0.3 and +0.1 respectively from previous best experiments. This experiment shows while HintedBT does benefit from more data, the increase in performance does not commensurate to the large increase in volume of data.  
 
%We test this by running We report results in Table \ref{tab:exps_more_data}. Rows 1, 2 represent experiments with 23M BT data, and rows 3, 4 with 46M BT data.

% We operated within a bounded setting - what if we used more data? We experiment with double the amount of monolingual data as before. Results in Table \ref{tab:exps_more_data}.

%We see that in most cases, the addition of data helps the model improve further. The performance in hi$\rightarrow$en increases further to a score of \textbf{32.0} when quality tags are used with 46M BT data. Performance in gu$\rightarrow$en improves to \textbf{18.2} when quality tags + translit tags are used with 46M BT data.  We see that here as well, topk-BT is the best result for ta$\rightarrow$en. We discuss this further in Section \ref{tab:ta2en_analysis_exps}. {\color{red}{should we make a comment about ta2en here? it is still less than topk with 23M}}

\begin{table}[h]
    \centering
    \resizebox{0.95\linewidth}{!}{
    \begin{tabular}{|c|l|l|l|l|}
    \hline
    $\#$ & \textbf{Data} & \textbf{hi$\rightarrow$en} & \textbf{gu$\rightarrow$en} & \textbf{ta$\rightarrow$en} \\ \hline
    & \textbf{Using 23M BT} & - & - & - \\
    \hline
    1 & bitext + full-BT & 30.9 & 15.0 & 14.1 \\ \hline
    2 & Row-1 + LaBSE qual.tags & 31.2 & 17.6 & 15.5 \\ \hline
    3 & Row-2 + Translit-tags & 31.6 & 17.9 & 16.0 \\ \hline
    % 3 & \textit{Jacc.} qual.tags & 31.3 & 17.8 & 15.7 \\ \hline
    % 4 & + Translit-tags & 31 & 17.7 & \textbf{16.3} \\ \hline
    & \textbf{Using 46M BT} & - & - & - \\ \hline
    4 & bitext + full-BT & 31.3 & 14.9 & 14.4 \\ \hline
    5 & Row-4 + LaBSE qual.tags & \textbf{32.0} & 17.9 & 16.0 \\ \hline
    6 & Row-5 + Translit-tags & 31.3 & \textbf{18.2} & \textbf{16.1}\\ \hline
    % 7 & \textit{Jacc.} qual.tags & 31.4  & 17.6 & 16.1 \\ \hline
    % 8 & + Translit-tags & 31.6 & 17.7 & 15.6\\ \hline
    % LaBSE qual.tags & \textbf{32}(+0.8) & 17.9(+0.3) & 16(+0.5) \\ \hline
    % + Translit-tags & 31.3(-0.3) & \textbf{18.2}(+0.3) & 16.1(+0.1)\\ \hline
    % \textit{Jacc.} qual.tags & 31.4(+0.1)  & 17.6(-0.2) & 16.1(+0.6) \\ \hline
    % + Translit-tags & 31.6(+0.6) & 17.7(+0) & 15.6(-0.7)\\ \hline
    \end{tabular}}
    \caption{Experiments with a larger monolingual corpus. Rows 4-6 are directly comparable to rows 1-3.}
    \label{tab:exps_more_data}
\end{table}

\section{Conclusion}
In this work, we propose HintedBT, a family of techniques that adds hints to back-translation data to improve their effectiveness. We first propose \textit{Quality Tagging} wherein we add tags to the source which indicate the quality of the source-target pair. We then propose \textit{Translit Tagging} which uses tags on the target side corresponding to the translation/transliteration operations that are required on the source. We present strong experimental results over competitive baselines and demonstrate that models trained with our tagged data are competent with state-of-the-art systems for all three language pairs. The application of our techniques to multilingual models and to other generation techniques for back-translation (such as noised beam \cite{edunov2018understanding}) are interesting avenues for future work.
% To produce a PDF file, pdf\LaTeX{} is strongly recommended (over original \LaTeX{} plus dvips+ps2pdf or dvipdf). Xe\LaTeX{} also produces PDF files, and is especially suitable for text in non-Latin scripts.

\section*{Acknowledgements}
We thank Gustavo Hernandez Abrego, Jason Riesa, Julia Kreutzer, Macduff Hughes, Preksha Nema, Sneha Mondal and Wolfgang Macherey for their early reviews and helpful feedback. We also thank the reviewers for their valuable and constructive suggestions.

\bibliography{anthology,custom}
\bibliographystyle{acl_natbib}

% \clearpage
\appendix

\section{Topk-BT Baseline} \label{sec:appendix_baseliens}
We report our extensive experiments for finding the best topk-BT data here for both LaBSE and BoT-Jaccard based scoring of back-translated data pairs.

\begin{table}[h]
    \centering
    \resizebox{0.95\linewidth}{!}{
    \begin{tabular}{|c|c|c|}
    \hline
    \textbf{Data used} & \textbf{WMT test set} & \textbf{Dev set} \\ \hline \hline
    LaBSE top 6.5M & 29.6 & 20.1 \\ 
    LaBSE top 8M & 30.4 & 20.2 \\
    LaBSE top 10M & 30.6 & 20.5 \\
    LaBSE top 15M & 30.7 & 20.3 \\
    LaBSE top 18M & \textbf{31.2} & 20.3 \\
    LaBSE top 20M & \textbf{31.2} & \textbf{20.6} \\
    Full-BT & 30.9 & 20 \\ \hline \hline
    BoT-Jaccard top 6.5M & 30.4 & 20.8 \\ 
    BoT-Jaccard top 8M & 30.7 & \textbf{21.1} \\
    BoT-Jaccard top 10M & 30.4 & 20.5 \\
    BoT-Jaccard top 15M & \textbf{31} & 20.6 \\
    BoT-Jaccard top 18M & 30.6 & 20.5 \\
    BoT-Jaccard top 20M & 30.7 & 20.3 \\
    Full-BT & 30.9 & 20 \\ \hline
    \end{tabular}}
    \caption{Grid search for best topk-BT data for hi$\rightarrow$en}
    \label{tab:fqe_hi2en}
\end{table}

\begin{table}[h]
    \centering
    \resizebox{0.95\linewidth}{!}{
    \begin{tabular}{|c|c|c|}
    \hline
    \textbf{Data used} & \textbf{WMT test set} & \textbf{Dev set} \\ \hline \hline
    LaBSE top 650k & 13 & 22.3 \\ 
    LaBSE top 1M & 13 & 23.3 \\
    LaBSE top 3.5M & 15.4 & 26.3 \\
    LaBSE top 6.5M & 15.7 & 27 \\
    LaBSE top 8M & \textbf{16.3} & 26.9 \\
    LaBSE top 10M & 16 & \textbf{27.2} \\
    LaBSE top 15M & 16.1 & 26.8 \\
    Full-BT & 15 & 25.8 \\ \hline \hline
    BoT-Jaccard top 650k & 12.2 & 21.5 \\ 
    BoT-Jaccard top 1M & 12.8 & 22.3 \\
    BoT-Jaccard top 3.5M & 15 & 25.7 \\
    BoT-Jaccard top 6.5M & 15.4 & 26.7 \\
    BoT-Jaccard top 8M & 16 & 26.9 \\
    BoT-Jaccard top 10M & \textbf{16.3} & \textbf{27} \\
    BoT-Jaccard top 15M & 15.5 & 26.7 \\
    Full-BT & 15 & 25.8 \\ \hline
    \end{tabular}}
    \caption{Grid search for best topk-BT data for gu$\rightarrow$en}
    \label{tab:fqe_gu2en}
\end{table}

\begin{table}[h]
    \centering
    \resizebox{0.95\linewidth}{!}{
    \begin{tabular}{|c|c|c|}
    \hline
    \textbf{Data used} & \textbf{WMT test set} & \textbf{Dev set} \\ \hline \hline
    LaBSE top 2.5M & 15.5 & 19 \\ 
    LaBSE top 5M & \textbf{16.4} & \textbf{19.9} \\
    LaBSE top 8M & 16 & 19.8 \\
    LaBSE top 10M & 15.5 & 18.6 \\
    LaBSE top 15M & 15.5 & 19.3 \\
    LaBSE top 20M & 14.6 & 18.7 \\
    Full-BT & 14.1 & 18.5 \\ \hline \hline
    BoT-Jaccard top 2.5M & 15.1 & 18.8 \\
    BoT-Jaccard top 5M & \textbf{16.5} & \textbf{19.6} \\
    BoT-Jaccard top 8M & 15.4 & 19.3 \\
    BoT-Jaccard top 10M & 15.1 & 19.3 \\
    BoT-Jaccard top 15M & 15.1 & 18.6 \\
    Full-BT & 14.1 & 18.5 \\ \hline
    \end{tabular}}
    \caption{Grid search for best topk-BT data for ta$\rightarrow$en}
    \label{tab:fqe_ta2en}
\end{table}

\section{Experimental Setup} \label{app:exp_setup_app}
We experiment with the following hyper-parameters - 
\vspace{-2mm}
\begin{itemize} [noitemsep]
    \item Number of encoder-decoder layers - 4, 6
    \item Number of attention heads - 12, 16
    \item Embedding dimensions - 768, 1024
    \item Hidden dimension - 1536, 8192
\end{itemize}
We choose the final model configuration described in Section \ref{subsec:Model Architecture} based on the dev-BLEU scores of the respective bitext models. 
% We saw that gu$\rightarrow$en required a smaller model (4 encoder-decoder layers as compared to 6 for hi$\rightarrow$en), possibly because of the smaller amount of data available. 
However, further reduction of model size (reducing the number of attention heads, hidden dimension etc.) caused the models to underfit. The hi$\rightarrow$en models have 375M parameters and gu$\rightarrow$en and ta$\rightarrow$en models have 283M parameters. Training was done using Tensorflow-Lingvo \cite{shen2019lingvo}.

Note: For gu$\rightarrow$en and ta$\rightarrow$en, we randomly pick 200 pairs from each train source (from Table \ref{tab:datasets}) and append them to the WMT newsdev set for better diversity.\\

\section{Human Evaluation of Data Quality}
\label{app:human_eval}
We ask human raters to evaluate the quality of source-target pairs (similar to \citet{wu2016google}). 
Quality scores range from 0 to 6, with a score of 0 meaning ``completely nonsense translation'',and a score of 6 meaning ``perfect translation: the meaning of the translation is completely consistent with the source, and the grammar is correct''. A translation is given a score of 4 if ``the sentence retains most of the meaning of the source sentence, but may have some grammar mistakes'', and a translation is given a score of 2 if ``the sentence preserves some of the meaning of the source sentence but misses significant parts''.  These scores are generated by human raters who are fluent in both source and target languages.  \\

\noindent The final human evaluation score of a set of $n$ examples is given by the average of the $n$ individual scores. When comparing two systems side-by-side, the difference between their two final scores quantifies the change in quality. In this case, a difference of $\pm 0.1$ is considered significant. 
% The final ratings for both systems are then averaged and the difference between these two averages quantifies the change in quality. A difference of \textbf{$\pm$0.1} is considered significant.

\section{FST Transliteration Models} \label{app:fst_translit_models}
To generate source to target language transliterations for \textit{Translit Tagging}, we use FST transliteration models from \citet{hellsten2017transliterated}. Weighted Finite State Transducer (WFST) models are trained on individual word transliterations of native words from a set vocabulary, collected from 5 speakers amongst a large pool of speakers. These models are evaluated on annotated test sets for Hindi and Tamil, and they achieve 84\% and 78\% word-level accuracies respectively. 

\section{Number of Bins : Quality Binning} \label{app:bin_count}
We experiment with different number of bins in Equal Volume Binning for \textit{Quality Tagging}. We show our experiments and corresponding dev-BLEU scores in Table \ref{tab:bin_count_exps}. 
\begin{table}[h]
    \centering
    \resizebox{0.95\linewidth}{!}{
    \begin{tabular}{|l|l|l|l|l|}
    \hline
    \textbf{Data} & \textbf{hi$\rightarrow$en} & \textbf{gu$\rightarrow$en} & \textbf{ta$\rightarrow$en} \\ \hline 
    bitext + full-BT & 20.0 & 25.8 & 18.5\\ \hline
     + 3 LaBSE qual. tags & 20.6 & 28.0 & 18.4 \\
     + 4 LaBSE qual. tags & \textbf{20.8} & \textbf{28.4} & \textbf{18.6} \\
     + 5 LaBSE qual. tags & 20.5 & 28.0 & 18.2 \\
   %  + 5 LaBSE qual. tags & 31.3 & 17.6 & 15.4 \\
    \hline
    \end{tabular}}
    \caption{Quality-Tagging Experiments with different number of bins}
    \label{tab:bin_count_exps}
\end{table}

\section{Translit-tagging on the source side} \label{app:source_translit_tagging}
In previous sections, we trained models with translit tags on the target side, hence enabling the models to predict whether or not transliteration should be done on the source. An alternative method is to provide these translit tags as \textit{information} to the model, on the source side.

\noindent As we explain in Section \ref{sec:translit_tag_explain}, we require the target sentence to determine the translit tags. This is fine in the target-tagging case, since we do have access to the target while training; during inference, the model predicts the tag by itself. However, when we train a model with these tags on the \textit{source}, it becomes necessary to provide this tag during inference as well - this renders this method is infeasible at test time. 
We conduct an oracle experiment where we assume the right tags are available from the target at test time. We report the results in Table \ref{tab:seg_tag_base}. We see that for hi$\rightarrow$en and ta$\rightarrow$en, source tagging improves upon the full-BT baseline by +0.3; however for gu$\rightarrow$en source tagging is worse by -0.2. For hi$\rightarrow$en, source-tagging is better than target-tagging by +0.2; however for gu$\rightarrow$en and ta$\rightarrow$en, target-tagging is significantly better.

%still run and report experiments with translit tags on the source side, in the hypothetical case that we already have the necessary information from the target (an oracle case)
% Providing the translit tags as information to the model as tags on the source side, as opposed to training the model to predict  - this method depends upon the target to get the tag, which is okay while training, but not feasible at test time. We run these experiments even so, assuming that we have the tags at inference - but the results for gu2en and ta2en are still worse than having the model predict the tags, and results for hi2en are only slightly more significant than target tagging.

\begin{table}[h]
    \centering
    \resizebox{0.95\linewidth}{!}{
    \begin{tabular}{|l|l|l|l|}
    \hline
    \textbf{Data} & \textbf{hi$\rightarrow$en} & \textbf{gu$\rightarrow$en} & \textbf{ta$\rightarrow$en} \\ \hline
    bitext + full-BT & 30.9 & 15.0 & 14.1\\ \hline 
    \textbf{\textit{(+ translit tags)}} & - & - & - \\ \hline
    Source-Tagged & 31.2 & 14.8 & 14.4\\
    Target-Tagged & 31.0 & 15.2 & 15.0\\ \hline
    \end{tabular}}
    \caption{Source side Translit-tagging on full-BT data}
    \label{tab:seg_tag_base}
\end{table}

\section{Alternate Experiments for hi$\rightarrow$en}
In our hi$\rightarrow$en experiments, we use the IIT Bombay en-hi Corpus \cite{kunchukuttan2017iit} with 1.68M source-target pairs as the training dataset. In this section, we repeat our HintedBT experiments with the original training set from WMT-2014, which has 271k source-target pairs. We report test scores on the WMT-2014 hi$\rightarrow$en newstest set in Table \ref{tab:hien_origwmt}.

\begin{table}[h]
    \small
    \centering
    \resizebox{0.97\linewidth}{!}{
    \begin{tabular}{|l|l|}
    \hline
    \textbf{Modeling Methodology} & \textbf{Test} \\ \hline
    bitext & 10.3 \\
    bitext + full-BT & 25.5\\
    bitext + full-BT + \textit{LaBSE} quality tags & \textbf{27.4}  \\
    bitext + full-BT + \textit{LaBSE} quality tags + translit-tags & 27.0  \\ \hline
    \end{tabular}}
    \caption{Experiments with hi$\rightarrow$en WMT-2014 train set.}
    \label{tab:hien_origwmt}
\end{table}

\section{Comparison of BoT-Jaccard against LaBSE as a Quality Metric} \label{app:jacc_labse_qual_metric}
We run all the experiments in 
Section \ref{sec:exp_setup} with BoT-Jaccard scores in the place of LaBSE scores. We present results in Table \ref{tab:jaccard_exps}. We see for hi$\rightarrow$en, the topk-BT baseline is lower than the full-BT baseline, whereas for gu/ta$\rightarrow$en, topk-BT is higher. For hi/gu$\rightarrow$en, the BoT-Jaccard score based quality tagging gives competent results, whereas for ta$\rightarrow$en, the topk-BT model remains the best result.

\begin{table}[h]
    \centering
    \resizebox{0.95\linewidth}{!}{
    \begin{tabular}{|l|l|l|l|}
    \hline
    \textbf{Data} & \textbf{hi$\rightarrow$en} & \textbf{gu$\rightarrow$en} & \textbf{ta$\rightarrow$en} \\ \hline
    bitext & 19.5 & 8.4 & 11.3 \\ \hline
    bitext + full-BT & 30.9 & 15.0 & 14.1\\ \hline
    bitext + \textit{Jacc.} topk-BT & 30.7 & 16.3 & \textbf{16.5} \\ \hline
    bitext + full-BT    & \multirow{2}{*}{\textbf{31.3}} & \multirow{2}{*}{\textbf{17.8}} & \multirow{2}{*}{15.7} \\
    + \textit{Jacc.} quality tags & & & \\ \hline
    bitext + \textit{Jacc.} topk-BT & \multirow{2}{*}{30.7} & \multirow{2}{*}{16.1 } & \multirow{2}{*}{15.8} \\ 
    + translit-tags & & & \\ \hline
    bitext + full-BT  & \multirow{3}{*}{31.0} & \multirow{3}{*}{17.7} & \multirow{3}{*}{16.3} \\
     + \textit{Jacc.} quality tags  & & &  \\
    + translit-tags & & & \\ \hline
    \end{tabular}}
    \caption{Performance of models on WMT test sets, using BoT-Jaccard scoring. These results are directly comparable to corresponding rows in Table \ref{tab:main_results}.}
    \label{tab:jaccard_exps}
\end{table}

\begin{table}[h]
    \centering
    \resizebox{0.95\linewidth}{!}{
    \begin{tabular}{|l|l|l|l|}
    \hline
    \textbf{Quality Metric} & \textbf{hi$\rightarrow$en} &
    \textbf{gu$\rightarrow$en} & \textbf{ta$\rightarrow$en} \\ \hline
    BoT-Jaccard & 0.127 & 0.306 & 0.245  \\
    LaBSE & 0.262 & 0.399 & 0.314 \\ \hline
    \end{tabular}}
    \caption{Spearman's correlation coefficient for quality metrics against human judgements of quality (1500 samples for each). p-value < 0.001 for all scores.}
    \label{tab:metric_corr}
\end{table}

\begin{figure*}
    \centering
    \frame{\includegraphics[width=0.97\linewidth]{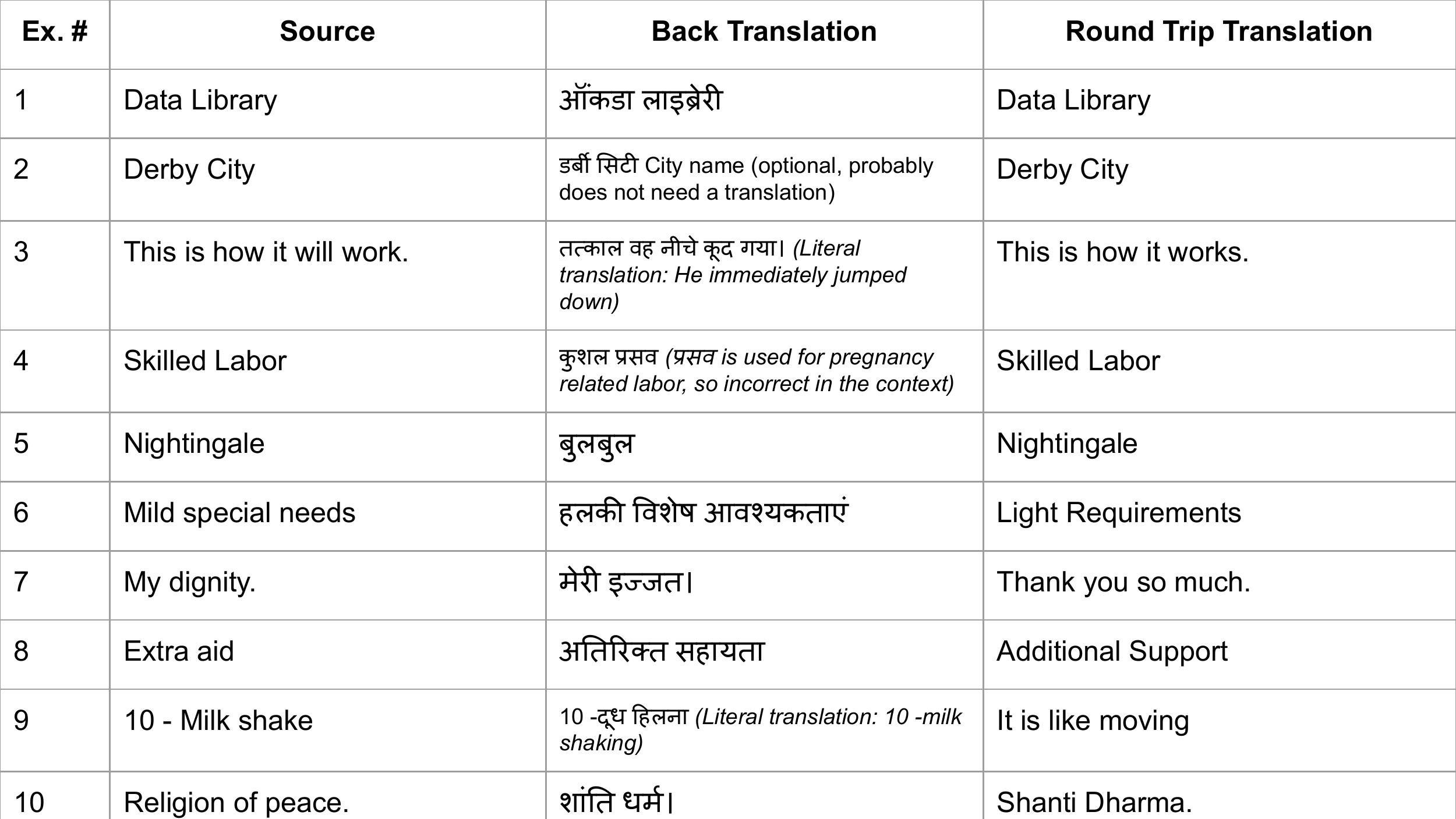}}
    \caption{Examples from the qualitative analysis of Jaccard and LaBSE scores. Text in italics are added as comments. Everything else is part of system output.}
    \label{fig:qual_examples}
\end{figure*}

\noindent To better understand patterns of LaBSE or BoT-Jaccard mistakes in evaluating quality for parallel data, we manually annotate back-translations for hi$\rightarrow$en where the metrics oppose each other. We select 200 random instances where, 
\begin{equation*}
\begin{split}
& abs(\text{BoT-Jaccard} - \text{LaBSE}) > 0.2 \\
\text{and }  &    min(\text{BoT-Jaccard}, \text{LaBSE}) < 0.5
\end{split}
\end{equation*}
We manually annotate which metric is correct, and the reason for the other metric’s failure. We present the analysis in two parts, one where BoT-Jaccard score is higher than LaBSE, and the other where LaBSE is higher than BoT-Jaccard. In Table~\ref{tab:qual_analysis_labse_ge_jac} and Table~\ref{tab:qual_analysis_jac_ge_labse} we present the categorizations of mistakes made by either method. Figure~\ref{fig:qual_examples} shows examples of source sentences, their back translations, and round trip translations which are referred to in the analysis.

\begin{table*}[h]
    % \small
    \centering
    \begin{tabular}{|p{0.22\linewidth}|p{0.03\linewidth}|p{0.65\linewidth}|}
    \hline
    \textbf{Reason} & \textbf{\#} & \textbf{Explanation} \\ \hline
    LaBSE cannot capture similarity for transliterations & 45 & This is probably because LaBSE has not been trained on parallel data which contains transliterations. Entities in particular are often transliterated (transcribed in Devanagari), but sometimes even common words like “library” are used through transliteration rather than translation, in Hindi. Row 1 in Figure~\ref{fig:qual_examples} is an example for this. \\ \hline
    
    Mistake in BT fixed by RTT deceives Jaccard & 40 & There are further three categories of mistakes in BT here.
    \begin{enumerate}
        \item The first is some random noise added to the BT probably stemming from the training data. The phrase "City name (optional, probably does not need a translation)" on Row 2 in Figure~\ref{fig:qual_examples} is generated by the BT model and corrected by the RTT model.
        \item Second, is a completely irrelevant BT that is somehow corrected by the RTT like Row 3 in Figure~\ref{fig:qual_examples}. This might also be due to faulty training data for the BT and RTT models.
        \item The last mode of failure is where the BT is wrong because it uses a wrong synonym for translating a source word like Row 4 in Figure~\ref{fig:qual_examples} where the word for pregnancy labor is used for translating the phrase ``skilled labor".
    \end{enumerate} \\ \hline
    
    LaBSE misses semantic similarity in source and BT & 15 & This is the least common mode of failure and might point to some gaps in LaBSE training for this language pair (not trained with enough data to cover rare synonyms or formulations). On Row 5 in Figure~\ref{fig:qual_examples}, LaBSE does not recognize the correct translation for ``Nightingale".\\
    \hline
    \end{tabular}
    \caption{Categorization and number of examples where LaBSE $>>$ Jaccard.}
    \label{tab:qual_analysis_labse_ge_jac}
\end{table*}

\begin{table*}[h]
    % \small
    \centering
    \resizebox{0.95\linewidth}{!}{
    \begin{tabular}{|p{0.22\linewidth}|p{0.03\linewidth}|p{0.65\linewidth}|}
    \hline
    \textbf{Reason} & \textbf{\#} & \textbf{Explanation} \\ \hline
    Model translating BT to RTT makes a mistake and deceives Jaccard & 46 & In the most common case, the Back Translation is correct, and this is correctly captured by LaBSE. However, the model translating BT to RTT makes a mistake, and therefore fools Jaccard on this instance. Row 6 in Figure~\ref{fig:qual_examples} is an example of slight difference in meaning between the correct BT and the RTT. In Row 7 the BT is correct, however the RTT is completely random. \\ \hline
    
    Synonyms used in RTT which preserves meaning but deceives Jaccard & 41 & In the second most common case, both the BT and RTT seem to have the same meaning as the original source sentence. However, the model translating BT to RTT uses synonyms of words in the source and therefore results in a low Jaccard score. Row 8 in Figure~\ref{fig:qual_examples} is an example of this. \\ \hline
    
    Mistake in both BT and RTT - wrongly marked as close by LaBSE & 9 & In this case, there is a slight mistake in meaning when source is translated to BT and it is further compounded by RTT. However, LaBSE marks the source and BT as close, which is incorrect. Row 9 in Figure~\ref{fig:qual_examples} is an example of this.\\ \hline
    
    Reverse model transliterates, which deceives Jaccard & 5 & Finally, in some examples, the reverse model transliterates the BT instead of translating it, resulting in low Jaccard scores. Row 10 in Figure~\ref{fig:qual_examples} is an example of this.\\
    
    \hline
    \end{tabular}}
    \caption{Categorization and number of examples where Jaccard $>>$ LaBSE.}
    \label{tab:qual_analysis_jac_ge_labse}
\end{table*}

\end{document}